%% file: main.tex
\documentclass[letterpaper]{article} % DO NOT CHANGE THIS
\usepackage{aaai2026}  % DO NOT CHANGE THIS
\nocopyright 
\usepackage{times}  % DO NOT CHANGE THIS
\usepackage{helvet}  % DO NOT CHANGE THIS
\usepackage{courier}  % DO NOT CHANGE THIS
\usepackage{booktabs}
\usepackage{multirow}
\usepackage[hyphens]{url}  % DO NOT CHANGE THIS
\usepackage{graphicx} % DO NOT CHANGE THIS
\usepackage{natbib}  % DO NOT CHANGE THIS AND DO NOT ADD ANY OPTIONS TO IT
\usepackage{caption} % DO NOT CHANGE THIS AND DO NOT ADD ANY OPTIONS TO IT
\usepackage{algpseudocode}
\usepackage{algorithm}
\usepackage{amsfonts} % 
\usepackage{amssymb} 
\usepackage{todonotes}
\usepackage{multicol}
\usepackage{graphicx}
\usepackage{tikz}
\usepackage[most]{tcolorbox}
\usepackage{lipsum} 
\usepackage{todonotes}
\usepackage{amsthm}
\newtheorem{definition}{Definition}
\usepackage{subcaption}
\usepackage{xcolor}
\definecolor{darkgreen}{rgb}{0.0, 0.5, 0.0}
\usetikzlibrary{shapes.geometric, arrows}

\tikzstyle{startstop} = [rectangle, rounded corners, minimum width=3cm, minimum height=1cm,text centered, draw=black, fill=red!30]
\tikzstyle{process} = [rectangle, minimum width=3cm, minimum height=1cm, text centered, draw=black, fill=orange!30]
\tikzstyle{decision} = [diamond, minimum width=3cm, minimum height=1cm, text centered, draw=black, fill=green!30]
\tikzstyle{arrow} = [thick,->,>=stealth]

\usepackage{newfloat}
\usepackage{listings}
\DeclareCaptionStyle{ruled}{labelfont=normalfont,labelsep=colon,strut=off} % DO NOT CHANGE THIS
\floatstyle{ruled}
\newfloat{listing}{tb}{lst}{}
\floatname{listing}{Listing}
\title{Adaptive Experiments Under Data Sparse Settings: Applications for Educational Platforms}
\author{
  Haochen Song\textsuperscript{1}, 
  Ilya Musabirov\textsuperscript{2}, 
  Ananya Bhattacharjee\textsuperscript{1},\\
  Audrey Durand\textsuperscript{3}, 
  Meredith Franklin\textsuperscript{1}, 
  Anna Rafferty\textsuperscript{4},
  Joseph Jay Williams\textsuperscript{1}\\ (
  \textsuperscript{1}University of Toronto, 
  \textsuperscript{2}University of British Columbia,
  \textsuperscript{3}Laval University, 
  \textsuperscript{4}Carleton College)
}

\usepackage{bibentry}
\begin{document}

\maketitle

\begin{abstract}
\input{sections/0_abstract}
\end{abstract}


\input{sections/1_introduction}
\input{sections/2_related_work}
\input{sections/3_high_dimensional}
\input{sections/4_method}
\input{sections/5_comparison}
\input{sections/6_problem_setting}
\input{sections/7_wapts_vs_ts}
\input{sections/8_lenient_regret}
\input{sections/9_discussion}

\bibliography{aaai2026}

\input{sections/appendix}

\end{document}

%% file: sections/0_abstract.tex
Adaptive experimentation is increasingly used in educational platforms to personalize learning through dynamic content and feedback. However, standard adaptive strategies such as Thompson Sampling often underperform in real-world educational settings where content variations are numerous and student participation is limited, resulting in sparse data. In particular, Thompson Sampling can lead to imbalanced content allocation and delayed convergence on which aspects of content are most effective for student learning.  To address these challenges, we introduce Weighted Allocation Probability Adjusted Thompson Sampling (WAPTS), an algorithm that refines the sampling strategy to improve content-related decision-making in data-sparse environments. WAPTS is guided by the principle of lenient regret, allowing near-optimal allocations to accelerate learning while still exploring promising content. We evaluate WAPTS in a learnersourcing scenario where students rate peer-generated learning materials, and demonstrate that it enables earlier and more reliable identification of promising treatments. 

%% file: sections/1_introduction.tex
\section{Introduction}

Adaptive learning is increasingly applied to personalize digital education experiences \cite{reza_mooclet_2021, ruan2024reinforcement}, including individualized learning paths \cite{shawky_towards_2019} and automated feedback and assessment \cite{abdelshiheed_leveraging_2023, st-hilaire_new_2022}. A widely-used framework for enabling such personalization is the multi-armed bandit (MAB) model \cite{lattimore_bandit_2020}, which supports sequential, data-driven decision making by learning from observed outcomes over time. For example, a bandit policy might identify the most effective explanation for a machine learning concept for a given student. Among bandit strategies\footnote{We use "policies” to refer to different treatment assignment policies throughout this paper.}, Thompson Sampling (TS) is widely recognized for its theoretical guarantees and empirical performance, particularly in settings where treatments must adapt to diverse learner profiles \cite{agrawal_analysis_2012}. Compared to confidence-based methods such as Upper Confidence Bound (UCB), TS is generally more robust to noisy feedback and delayed rewards, conditions commonly found in online learning environments \cite{chapelle_empirical_2011}.

Despite these advantages, TS and other adaptive strategies do not inherently solve a fundamental challenge in educational experimentation: the curse of dimensionality. As the number of possible treatments grows, the sample size required to distinguish between them increases substantially. While adaptive methods promise greater efficiency by allocating more samples to promising treatments and fewer to underperformers, they do not necessarily offer greater statistical confidence in selecting the optimal treatment. In fact, recent studies have shown that adaptive designs can underperform in small-sample educational contexts—sometimes requiring more data than uniform random assignment (UR) to achieve comparable levels of statistical certainty \cite{williams_challenges_2021, kaufmann_complexity_2016}.

To address these challenges, we introduce a new bandit policy: Weighted Allocation Probability Adjusted Thompson Sampling (WAPTS). This policy is designed to support two guiding questions:

\begin{itemize}
    \item \textbf{GQ1:} How can adaptive experiments operate effectively in a data-sparse educational settings?
    \item \textbf{GQ2:} How do we identify multiple near-optimal treatments while minimizing lenient regret?
\end{itemize}

WAPTS modified the standard TS sampling procedure by introducing a probabilistic adjustment mechanism to rebalances exploration and exploitation in data-sparse environments. The policy is grounded in the concept of lenient regret  \cite{merlis_lenient_2021}, which relaxes the traditional goal of finding the single best treatment. Instead, it prioritizes reliably identifying a subset of high-performing options— and approach more suited to real-world educational constraints.

This shift toward leniency acknowledges the practical realities  faced by education researchers and platform designers. When the number of observable interactions is limited, it is often more feasible to identify “good enough” treatments rather than making fine-grained distinctions that require high statistical power. WAPTS is designed to facilitate this type of decision-making by increasing the probability of converging on a robust set of near-optimal treatments.

We evaluate WAPTS using a case study on an online learnersourcing platform \cite{khosravi2023learnersourcing} by comparing bandit-based adaptive experimentation with traditional A/B testing in a data-sparse setting. The platform features a large pool of candidate treatments but offers limited evaluation bandwidth per user, reflecting a growing trend in education: an abundance of instructional options (including AI-generated content) paired with strict constraints on time, resources, and experimental sample size.

In simulation studies, WAPTS consistently outperforms traditional TS in key metrics, including 1) higher probability of the correctly selection of near-optimal treatment, 2) faster posterior convergence to near-optimal treatments, and 3) comparable stability in hypothesis testing outcomes. These results are most notable under data-sparse conditions and in environments with many competing treatments.

%% file: sections/2_related_work.tex
\section{Related Work}

Adaptive experimentation is increasingly applied in education to improve instruction by tailoring content in real time, which can enhance engagement and learning gains \cite{reza_mooclet_2021}. The \textbf{bandit framework} supports this process by balancing exploration of new treatments with exploitation of promising ones, making it suitable for personalized learning \cite{agrawal_analysis_2012, chapelle_empirical_2011}. A growing application is \textbf{learnersourcing}, where students generate and evaluate content such as examples or quiz questions \cite{khosravi2023learnersourcing}. Adaptive policys can prioritize high-quality contributions, but in small classes or low-participation MOOCs, strategies like TS and Upper Confidence Bound (UCB) may still require many interactions to discard suboptimal content \cite{williams_challenges_2021}.

While contextual bandits \cite{li_contextual-bandit_2010, tang_ensemble_2014} can personalize treatments, their sample complexity and feature demands often make them impractical at small scale. For tractability, we focus on standard (non-contextual) bandits, which remain common in adaptive educational systems \cite{dimakopoulou_online_2021}, though they struggle when treatment sets are large, data are sparse, or fine-grained choices are needed. Prior work reflects these challenges: \citet{doroudi_not_2019} used bandits for passive learnersourcing to identify effective explanations efficiently; \citet{mandel_where_2017} explored human-in-the-loop reinforcement learning, emphasizing limits of exploration with small samples; and \citet{kumar_using_2024} applied bandits to engagement tasks, noting trade-offs between adaptivity and statistical confidence.

In these settings, classic regret minimization—where any deviation from the best treatment is penalized—can be overly conservative \cite{feng_satisficing_2025}. Common mitigation techniques include model simplification and subgroup clustering. While simplification reduces computational burden, it often sacrifices fine-grained personalization~\cite{williams2016axis}. Clustering can help address data sparsity, but may fragment the sample pool further—particularly problematic in education contexts where tailoring learning resources to subpopulations (e.g., AI learners) is often critical \cite{4528962}.

In contrast, the \textit{lenient regret} framework tolerates limited suboptimal selections if it leads to faster, broader gains \cite{merlis_lenient_2021}. This perspective better reflects educational realities, where decision-makers often operate under strict resource ceilings and seek good-enough solutions over perfect ones. Our work builds on this insight, proposing WAPTS as a policy to allocate treatments more pragmatically in data-sparse environments.

%% file: sections/3_high_dimensional.tex
\section{Exploring Adaptive Policy Behavior in Data-Sparse Settings}

We evaluate outcomes in two ways: (1) the overall benefit to participants during the study, and (2) the post-study analysis to assess treatment effects. In classroom experiments, small class sizes and multiple treatments create data sparsity, limiting both discovery and average student benefit.

We define the following data-sparse situation as researchers might see in the experiment settings:

\textbf{Definition (data-sparse experiment):}
\begin{itemize}
    \item Sample size is too small to statistically distinguish between many treatments.
    \item Treatment effects are small or similar, making strict optimization inefficient.
\end{itemize}

In this setting, outcomes between treatments can vary substantially or be closely clustered—making it difficult to distinguish the truly best options from the rest. We will use later sections to explore how assumptions about this variability affect policy performance and design.

Traditional adaptive allocation policy such as Thompson Sampling often fail in such settings. They (1) assume strict regret minimization is feasible, and (2) may over-explore, wasting evaluations on indistinguishable options. In contrast, \textbf{lenient regret} allows some tolerance for near-optimal choices, prioritizing practical gains over theoretical optimality \cite{merlis_lenient_2021}.

To better understand the behavior of adaptive policies under data-sparse conditions, we conducted a series of simulations modeled under the following setup. For an experiment with $N$ participants and $K$ treatments, for simplicity, each treatment outcome has a binary outcome that follows a $Bernoulli(\theta_i)$ distribution. For simulation purposes, each treatment has a baseline success probability $\theta_{\min}$, and the optimal treatment has an increased success probability of $\theta_{\min} + \Delta$, where $\Delta$ is the \textbf{effect size}.

We evaluate two policies in parallel:
\begin{itemize}
    \item \textbf{Uniform Random (UR):} Students are assigned to treatments uniformly at random.
    \item \textbf{Thompson Sampling (TS):} Assignments are adaptively determined based on observed feedback.
\end{itemize}

\subsection*{Data-Sparse Tradeoffs}

\begin{figure}[h!]
    \centering
    \includegraphics[width=0.47\textwidth]{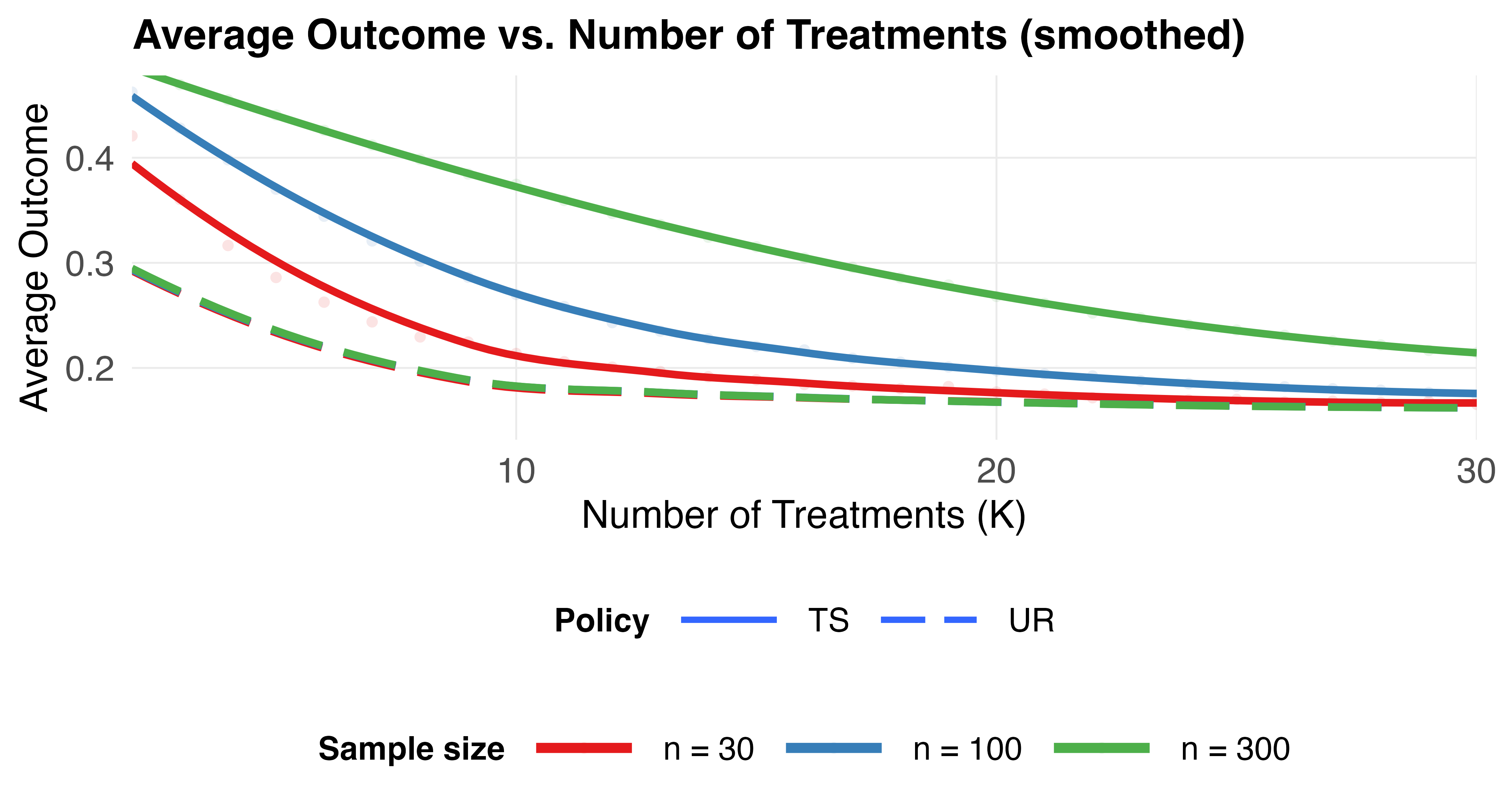}
    \caption{Average outcome vs. Number of Treatments (sample size $N=30, 100, 300$).}
    \label{fig1}
\end{figure}

Figure~\ref{fig1} shows average outcomes when sub-optimal treatments have success probabilities drawn i.i.d. from $\mathrm{Uniform}[\theta_{\min}, \theta_{\min}+\Delta]$ with $\theta_{\min}=0.15$ and $\Delta=0.35$ (so $\theta^*=\theta_{\min}+\Delta$). We consider three sample sizes, $N\in\{30,100,300\}$, and vary the number of treatments $K$ from 2 to 30. As $K$ increases, the average outcome declines for both UR and TS. With a small sample ($N=30$), TS shows little or no advantage over UR—there are too few observations per treatment (e.g., $N=K=30$) to explore effectively. For larger $N$ ($100$ and $300$), TS yields higher average outcomes, but its advantage shrinks as $K$ grows.

\subsection*{Posterior Estimates and Allocation Dynamics}

To further investigate why adaptive methods underperform in sparse settings, we examined how TS updates its beliefs and allocates participants across treatments. We picked  an experiment with 10 treatments, with all sub-optimal treatments fixed at $\theta_{\min} = 0.5$ and varying $\Delta$. Sample size is fixed at $N=1000$ to ensure convergence. 

\begin{figure}[h!]
    \centering
    \includegraphics[width=0.45\textwidth]{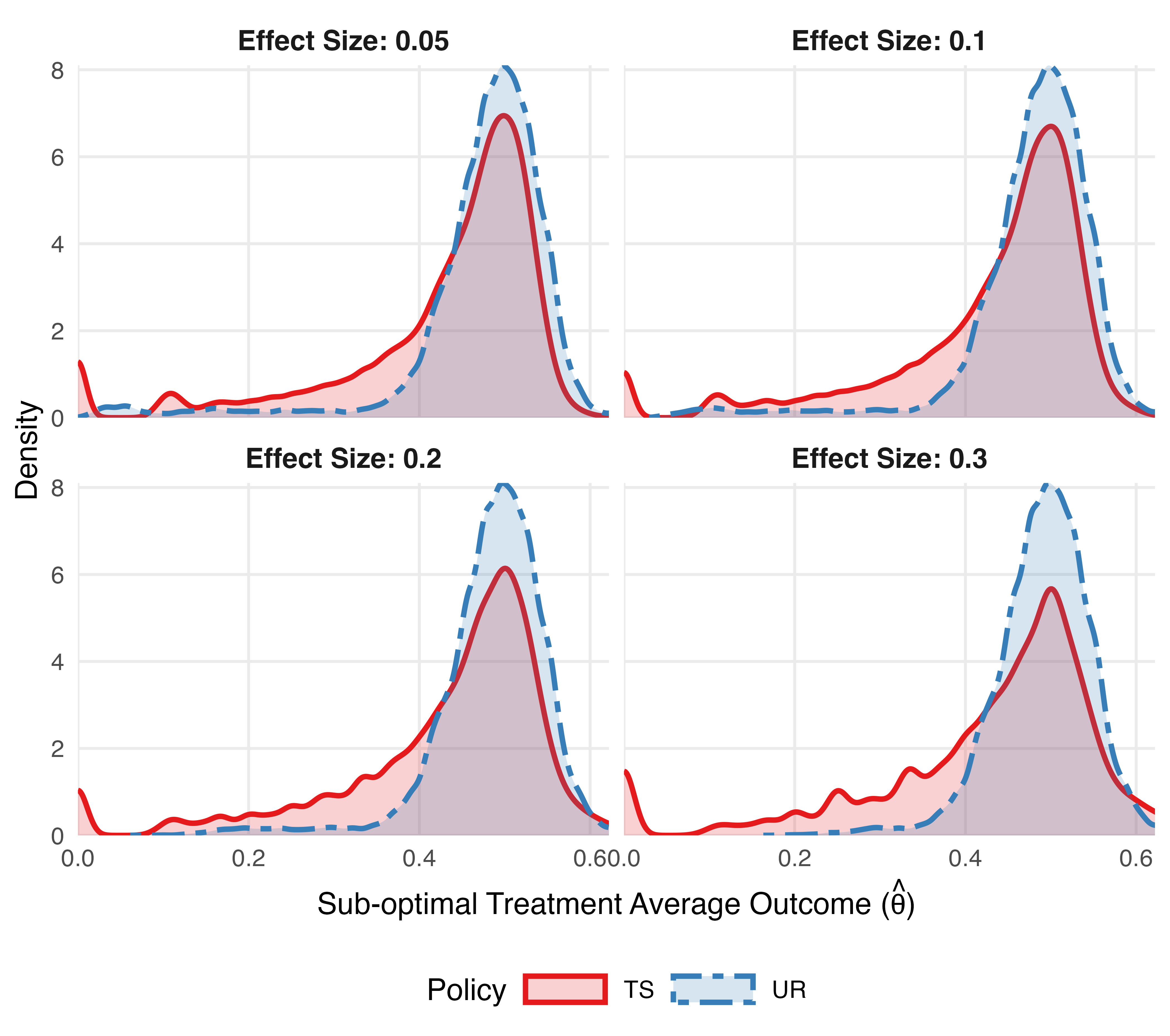}
    \caption{Posterior mean estimate for a sub-optimal treatment vs. effect size $\Delta$.}
    \label{fig:mean_estimate_policy}
\end{figure}

\begin{figure}[h!]
    \centering
    \includegraphics[width=0.45\textwidth]{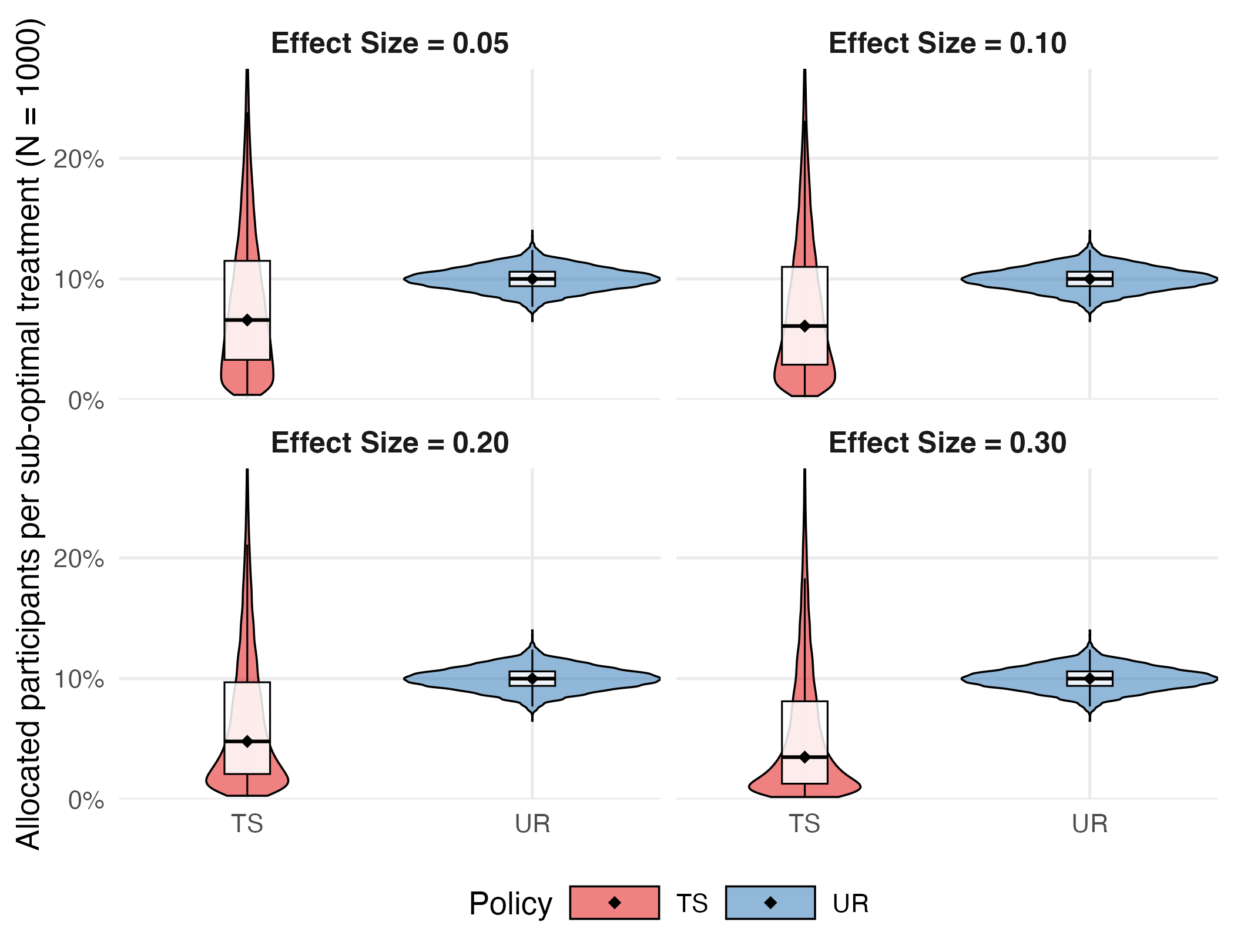}
    \caption{Violin plot of participant allocations in sub-optimal treatments as compared to effect size $\Delta$.}
    \label{fig:allocation_number}
\end{figure}

As shown in Figure~\ref{fig:mean_estimate_policy}, when $\Delta$ increases, posterior estimates of sub-optimal treatments becomes increasingly biased downward. This is due to early negative observations leading to premature de-allocation. Figure~\ref{fig:allocation_number} confirms this behavior: as the reward gap widens, TS assigns fewer participants to sub-optimal treatments, further reducing estimation quality. This effect has also been observed in prior work \cite{erraqabi_trading_2017, rafferty_statistical_2019}, even in lower-dimensional settings. 

In a typical classroom, the “treatments” are alternative worked examples, hints, or practice problems. In an adaptive system, an option that happens to receive a few weak ratings early is then shown to far fewer students. As a result, materials that might help specific subgroups (e.g., students with a particular misconception) are under-explored and under-estimated, and overall content diversity narrows.

%% file: sections/4_method.tex
\section{Method: Weighted Allocation Probability Adjusted Thompson Sampling (WAPTS)}
\label{sec:method}

Building from the diminishing behavior in Thompson Sampling we have shown in the data-sparse setting, we construct the idea of WAPTS policy, it is structured to work well in the data-sparse setting by early select a `nearly optimal or good' outcome treatment among the many treatments, so that these treatments could be selected exploitatively to make sure that there is not a huge loss. To show the performance of the policy, we further evaluate its performance a lenient regret setting.

\subsection*{Regret Formulations}

To evaluate the performance of WAPTS, we track both traditional and lenient regret:

\begin{definition}[Traditional Regret]
Let $\theta^* = \max\{\theta_1, \dots, \theta_K\}$ be the optimal success probability. If treatment $a_t$ is selected at trial $t$, the regret is
\[
r_t = \theta^* - \theta_{a_t}, \quad R_T = \sum_{t=1}^T r_t
\]
\end{definition}

\begin{definition}[Lenient Regret]
Given a tolerance $\epsilon > 0$, the lenient regret is:
\[
r_t^\epsilon = \mathbf{1}\{\theta^* - \theta_{a_t} > \epsilon\} \cdot (\theta^* - \theta_{a_t}), \quad R_T^\epsilon = \sum_{t=1}^T r_t^\epsilon
\]
\end{definition}

One insight here for experiment designers here is that leniency can be derived—not arbitrarily chosen—based on sample size, number of treatments, and effect estimates. This makes it better suited to educational domains where experimentation budgets are fixed.

\begin{itemize}
    \item Let $K$ denote the number of treatments from $Bernoulli(\theta_i)$ distribution for $i = 1,2,\dots,K$.
    \item There are a total of $N$ students who need to be assigned with different treatments.
    \item At trial $t$, a binary reward $r_t \in \{0,1\}$ is observed, which indicate if treatments are helpful based on learners' evaluations.
\end{itemize}

Each treatment $i$ has an unknown success probability $\theta_i$, modeled with a Beta posterior. Our method builds on standard Thompson Sampling (TS) by adjusting the sampling step to deprioritize treatments that are both empirically weak and uncertain. The resulting policy, WAPTS, aims to support experiments at a data-sparse setting, a full pseudo code can be found in the Appendix Section ~\ref{alg:wapts_full}, we also explain how the highlighted difference between WAPTS and TS:

\subsection*{Posterior Sampling with Weighted Allocation}

As shown in Algorithm~\ref{alg:wapts_full}, WAPTS differs from classical TS in the
\emph{treatment selection process}: instead of ranking arms by the Thompson draw alone, it
uses a weight that favors higher empirical success rates.Let $k_s[i]$ and $k_f[i]$ denote the counts of successes and failures for treatment $i$. We sample a posterior estimate $\tilde{p}_i$ for each treatment:

\[
\tilde{p}_i \sim \mathrm{Beta}(k_s[i] + 1,\; k_f[i] + 1)
\]

Then compute the empirical success rate:

\[
r_i = \frac{k_s[i]}{k_s[i] + k_f[i]}
\]

And finally define a weighted allocation score:

\[
\omega_i = (1 + r_i)\,r_i\,\tilde{p}_i
\]

This weighting factor tilts posterior sampling in favor of treatments that are empirically strong, helping to discard poor treatments earlier while preserving adaptive flexibility. A explore-exploit MAB policy such as TS will explore a sub-optimal treatment much more frequently due to the posterior belief of the policy, even though by looking at the posterior distribution, we would already know that such treatment is sub-optimal. Compared to TS, WAPTS is more efficient in resource-constrained settings with many treatments.

For a theoretical sketch of WAPTS’s regret bound and comparison with traditional Thompson Sampling, see Appendix~\ref{appendix:regret-wapts}. While WAPTS maintains the same posterior-sampling core as Thompson Sampling, its deterministic re-weighting can, in rare cases, downweight a truly optimal treatment to near-zero allocation due to early negative outcomes, preventing recovery and leading to unbounded regret. This does not occur in standard TS, where every treatment retains a nonzero sampling probability. To guarantee the theoretical $O(\log n)$ regret bound, an additional safeguard such as periodic forced exploration or allocation resets would be required.

%% file: sections/5_comparison.tex
\section{Comparison Between TS, UR, and WAPTS Across Different Sample Sizes}

We compare three allocation policies: UR, TS, and WAPTS. TS balances exploration and exploitation and guarantees an asymptotic convergence when there is infinite amount of samples. However, in practical settings with limited samples and many treatments, its convergence speed may be slow—requiring extended exploration to rule out suboptimal treatments. This can delay exploitation of the most effective options, which is costly in data-sparse educational experiments.

WAPTS addresses this limitation by accelerating the discovery of optimal treatments. It introduces a posterior reweighting mechanism that dynamically adjusts allocation probabilities based on observed empirical success rates. This refinement emphasizes treatments that perform better in early trials and penalizes poor performers, improving convergence speed without sacrificing adaptivity. This approach builds directly on the mechanism discussed in Section~\ref{sec:method}, where WAPTS penalizes low-performing treatments and boosts confidence in promising ones.

\begin{figure}[h]
    \centering
    \includegraphics[width=0.5\textwidth]{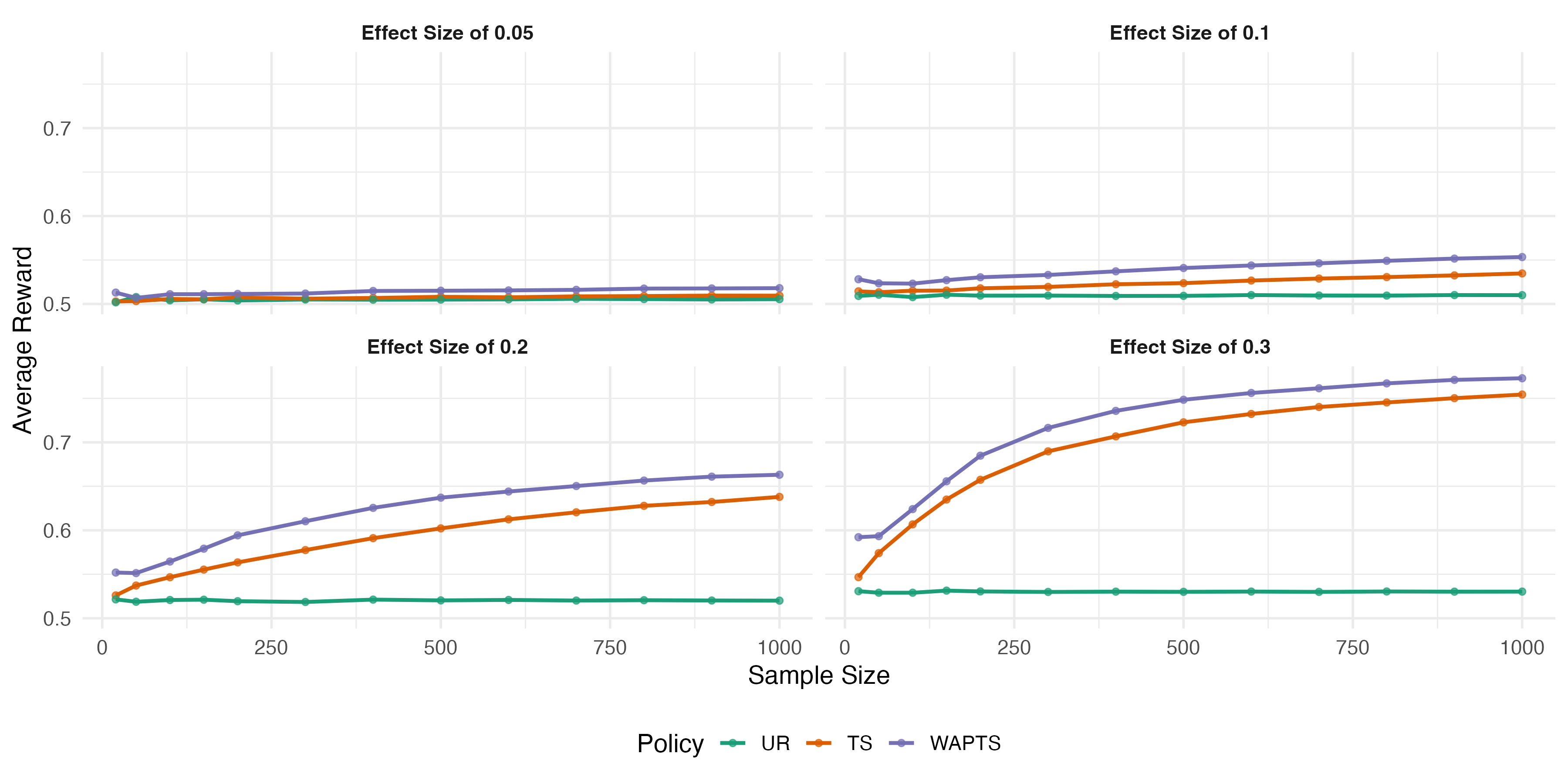}
    \caption{Policy comparison across varying sample sizes ($N$) and effect sizes ($\Delta$) under a 10-treatment setting. Results shown are average outcomes aggregated across 2000 replications per setting.}
    \label{fig:a2-outcome-all}
\end{figure}

We simulate 2000 hypothetical experiments under the \textbf{Fixed Gap Scenario} for each policy and each combination of effect sizes $\Delta \in \{0.05, 0.1, 0.2, 0.3\}$ and sample sizes ranging from 20 to 1000. As shown in Figure~\ref{fig:a2-outcome-all}, both adaptive methods outperform uniform random allocation across all conditions, though the magnitude of improvement depends on the effect size. Notably, WAPTS consistently achieves better average outcomes than traditional TS, especially at moderate to large effect sizes.

\begin{figure}[h]
    \centering
    \includegraphics[width=0.5\textwidth]{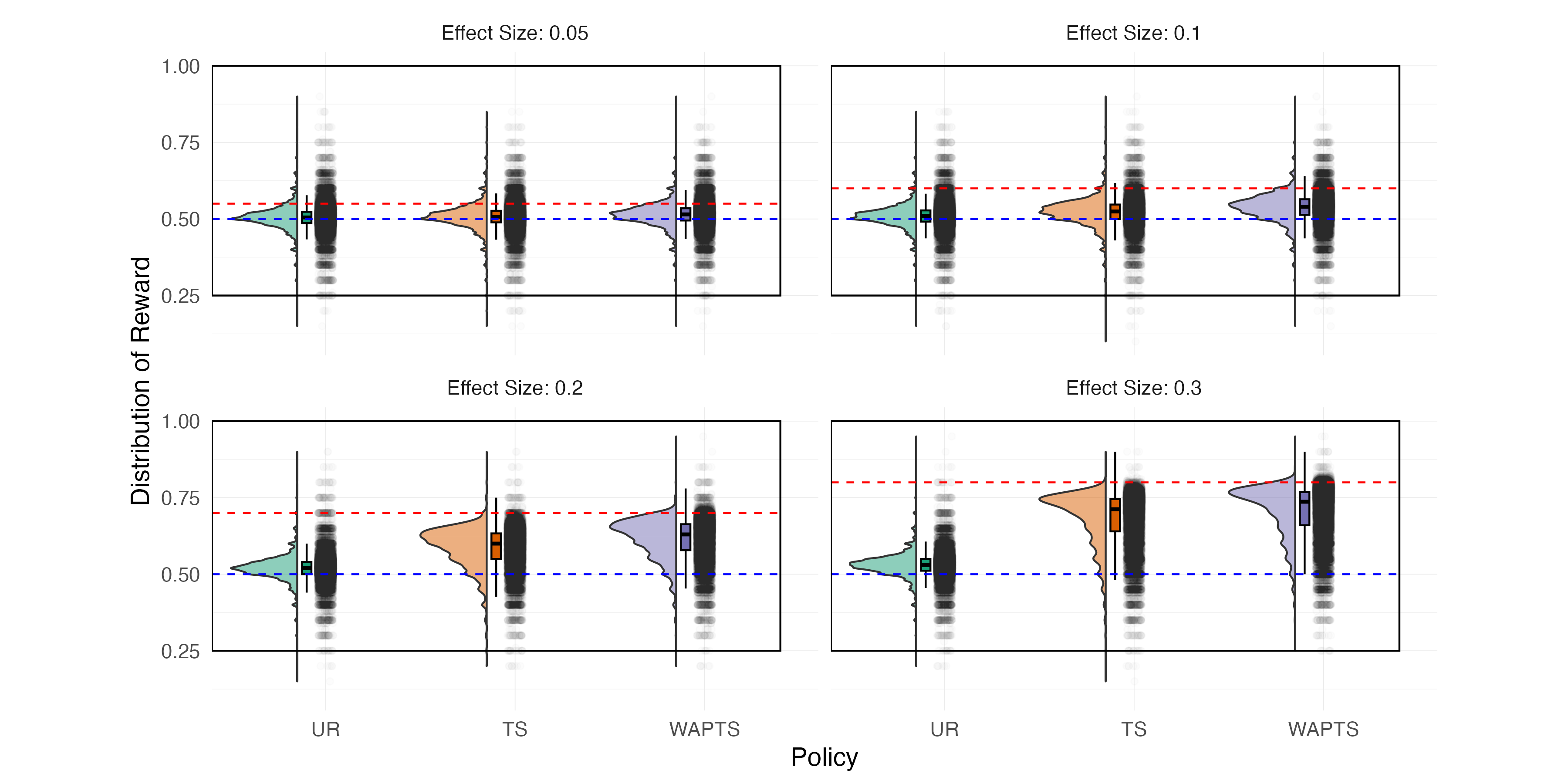}
    \caption{Distribution of average outcomes under each policy for a sample size of $N = 1000$ and varying effect sizes.}
    \label{fig:a2-outcome-1000}
\end{figure}

In learning design settings, sample sizes are often constrained by classroom enrollment. We therefore highlight the case of $N = 1000$, a realistic upper bound in our context. As shown in Figure~\ref{fig:a2-outcome-1000}, adaptive policys yield significantly better average performance, particularly when the effect size is nontrivial. For example, with $\Delta = 0.1$, WAPTS achieves an average outcome of 0.553 (SD = 0.025), compared with 0.535 (SD = 0.023) for TS and 0.510 (SD = 0.016) for UR. At $\Delta = 0.2$, the gap widens: WAPTS reaches 0.663 (SD = 0.027), TS at 0.638 (SD = 0.027), and UR remains at 0.520 (SD = 0.016).

These findings demonstrate the practical advantages of adaptive designs in education, particularly under real-world constraints. WAPTS offers a faster and more reliable path to identifying `near optimal' treatments, making it well-suited for settings where sample sizes are capped and immediate benefits to learners are prioritized. 

%% file: sections/6_problem_setting.tex
\section{Illustrative Scenario: Learnersourcing Under Resource Constraints}

We evaluate the performance of WAPTS by comparing it with Thompson Sampling (TS) and traditional A/B testing using a uniform random policy in an illustrative scenario of leveraging student examples (one kind of learnersourcing~\cite{khosravi2023learnersourcing}), first promoting generative learning~\cite{fiorella_eight_2016} and then participating in an active recall exercise \cite{ENDRES2024101974}.

In a university-level introductory statistics course, $N = 239$, each student proposed an example to explain a statistical concept (e.g., hypothesis testing or an adaptive experiment, see Appendix for examples of real student-generated explanations). 

Let us assume that after an initial screening to remove off-topic, trivial, or non-informative submissions, $K$ examples were selected for peer review. We describe how this peer review task can be formulated in a realistic resource-constrained setting and analyze how different policies perform. 

\begin{figure}[h]
    \centering
    \includegraphics[width=0.48\textwidth]{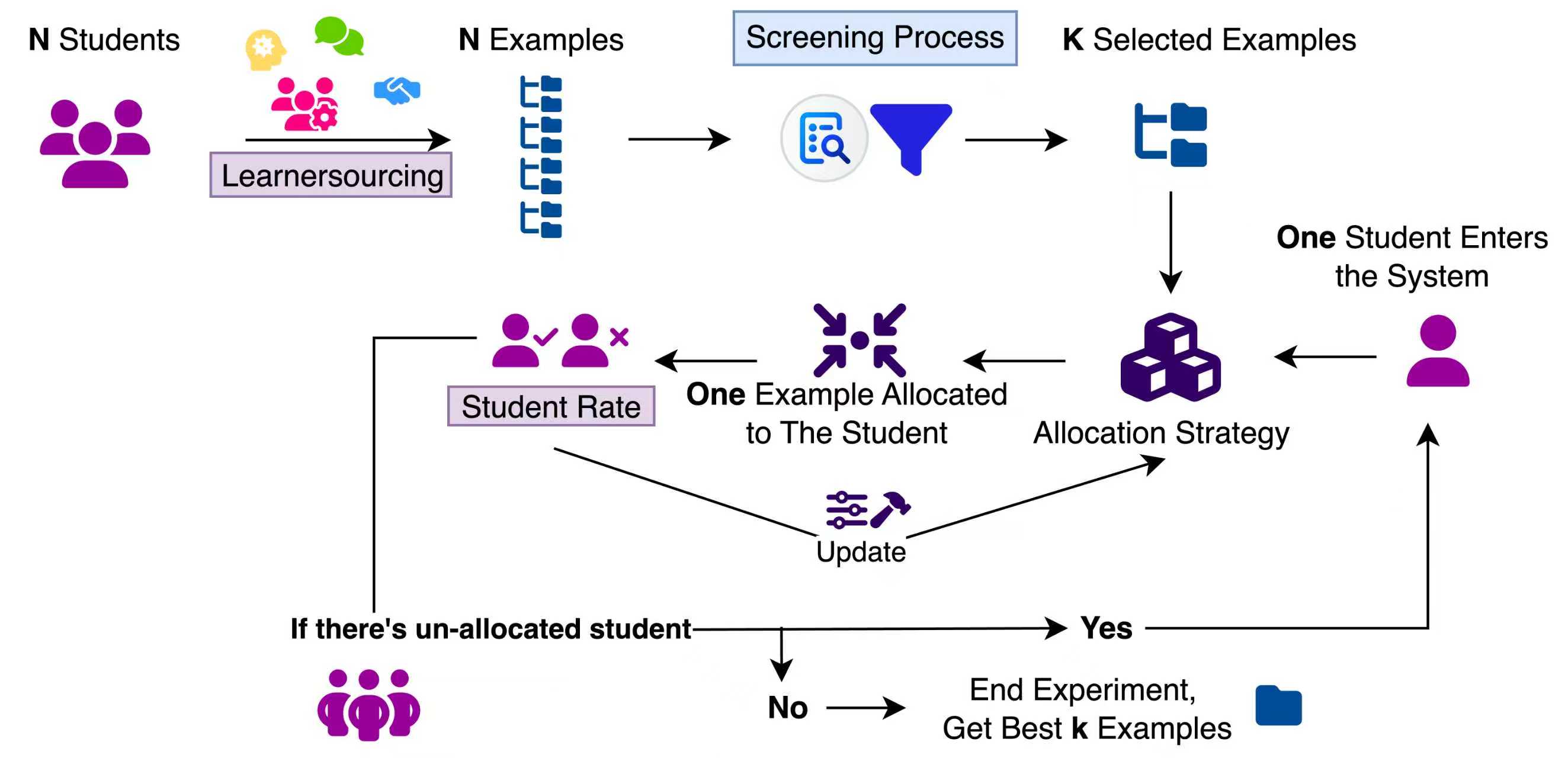} 
    \caption{Learnersourcing example flow: $N$ students create constructive retrieval examples. After an initial screening process, $K$ examples are selected and then passed forward for peer ratings.}
    \label{fig:flowchart}
\end{figure}

Due to time and attention constraints, each student can learn from and rate only one example out of the $K$ examples. Let us assume binary ratings (1 = helpful, 0 = unhelpful). The task can then be formulated as an adaptive allocation problem in a learnersourcing system (Figure \ref{fig:flowchart}): How can we identify and prioritize student engagement with high-quality examples with limited evaluations, assuming that several examples may be similarly instructive?

To formalize this problem, we assume $K \ll N$ examples (treatments), where each example $i$ has an unknown probability of being rated helpful, $\theta_i$. The highest such probability is $\theta^* = \max_{j} \theta_j$, and the goal is to identify all treatments that are (strictly or leniently) optimal.

We define a family of hypothesis tests, one per treatment:
\[
\begin{aligned}
H_{0,i}: &\ \theta_i \leq \theta^* - \epsilon \\
H_{1,i}: &\ \theta_i > \theta^* - \epsilon
\end{aligned}
\quad \text{for } i = 1, \ldots, K.
\]

\textbf{Estimation:}  
Since $\theta^*$ is unknown, we use its empirical estimate (e.g., the posterior mean from the bandit policy) for all comparisons. This introduces dependence across tests but reflects standard practice in adaptive decision making.

\textbf{Multiple Testing Interpretation:}  
Unlike global tests (which ask if \emph{any} treatment is optimal), our goal is treatment-level selection. We perform $K$ tests to determine whether each treatment meets the $\epsilon$-optimality criterion. This approach allows multiple treatments to pass the test and is practical in more data-sparse settings.

\textbf{Flexible Gaps and Lenient Optimality:}  
In some educational settings -- especially those involving accessible content or subjective preferences -- multiple treatments may be similarly effective. We model this using a flexible gap setting where treatment rewards are drawn from a uniform distribution:
\[
\theta_i \sim \mathrm{Uniform}(\theta_{\min}, \theta^*) \quad \text{with} \quad \theta^* = \theta_{\min} + \Delta.
\]
Here, $\theta_{\min}$ is the baseline expected effectiveness, and $\Delta$ controls the potential reward range. This allows multiple treatments to fall within an $\epsilon$-threshold of the best treatment, satisfying $\theta_i \geq \theta^* - \epsilon$.

In contrast, a fixed gap setting has exactly one optimal treatment with value $\theta^*$ and all others set to $\theta_{\min}$:
\[
\theta_i = \begin{cases}
\theta^*, & \text{for one } i \\
\theta_{\min}, & \text{for all other } i
\end{cases}
\quad \text{with} \quad \theta^* = \theta_{\min} + \Delta.
\]
In this case, lenient regret reduces to standard regret whenever $\epsilon < \Delta$.

In an ideal situation, the outcome of learnersourcing is to surface not just a single “best” artifact, but a small set of examples that are within an $\epsilon$-margin of the best. This aligns with the idea of lenient regret: if multiple artifacts are nearly as effective as the top one, then allocating attention across this subset still provides strong learning benefits. Having several $\epsilon$-optimal examples avoids overly relying on a single artifact, and exposes students to diverse content. 

On the other hand, achieving this with statistical guarantees requires assumptions about the distribution of good versus poor-quality submissions. If the pool contains many reasonably strong examples, then even as the number of treatments grows, lenient regret can remain small. In contrast, if only a few strong options exist among many poor ones, efficient identification becomes harder and regret is more difficult to control.

%% file: sections/7_wapts_vs_ts.tex
\section{Posterior Dynamics: Comparing WAPTS and Thompson Sampling}

To understand if WAPTS improves over standard Thompson Sampling (TS), we examine how the posterior estimates of each treatment evolve over time, especially under data-sparse conditions.

The WAPTS adjustment mechanism accelerates discard of low-performing treatments while boosting confidence in promising ones. For example, a treatment with 80\% success will be allocated with higher probability than one with only 20\%, due to the weighted adjustment factor $(1 + r_k) \cdot r_k$.

We compare posterior convergence across two reward structures: (1) a \textbf{Fixed-Gap} scenario where one optimal treatment stands out clearly, and (2) a \textbf{Flexible-Gap} scenario where several treatments are comparably good.

\subsection*{Posterior Estimates with a Sample Size of $N=239$}

\begin{figure}[h]
    \centering
    \includegraphics[width=0.5\textwidth]{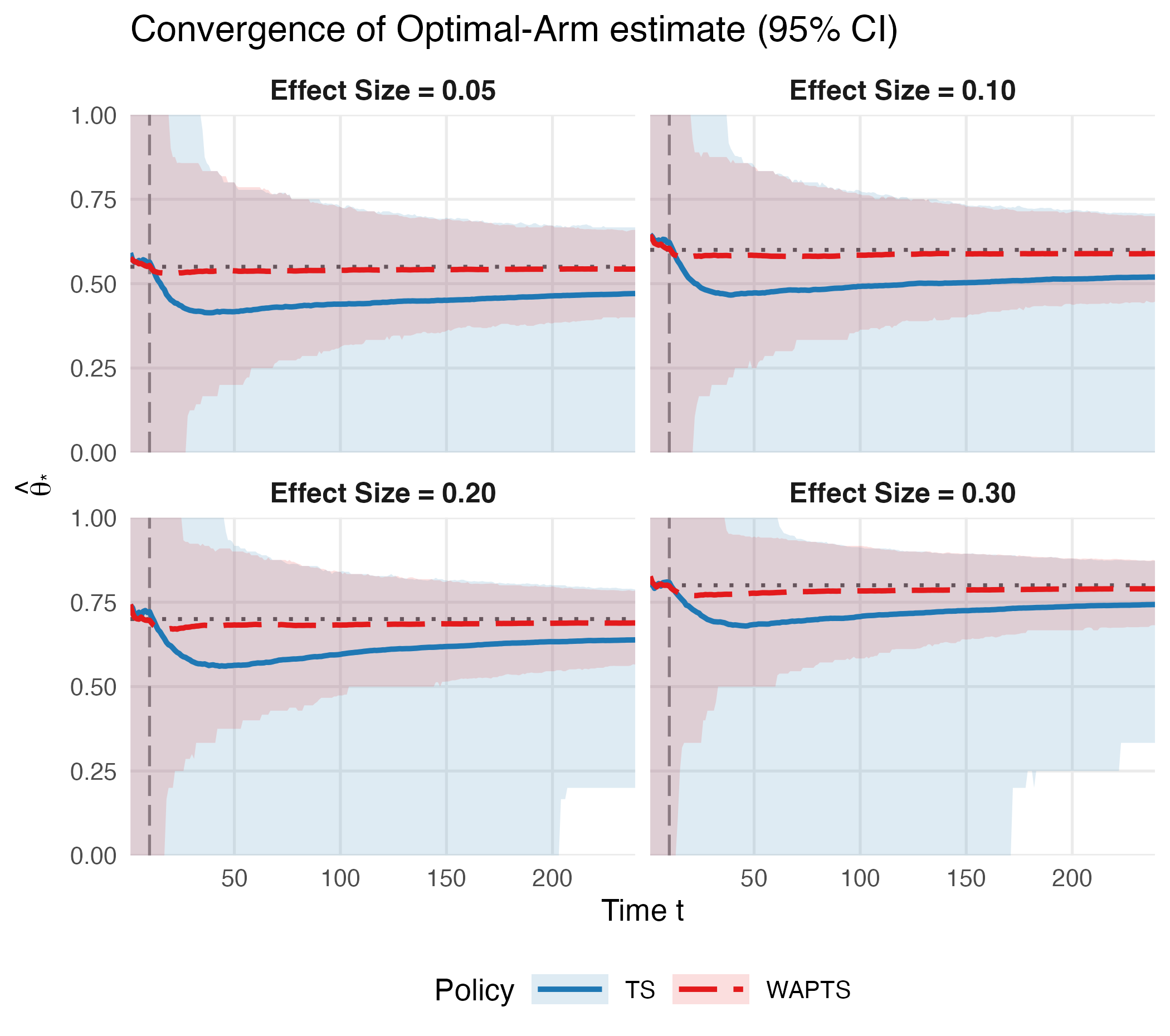}
    \caption{Posterior evolution with $N=239$ participants, shown under the flexible gap setting, with different effect size $\Delta$'s.}
    \label{fig:posterior_change_main}
\end{figure}

Figure~\ref{fig:posterior_change_main} summarizes posterior evolution for the flexible gap setting, under $N = 239$ with four effect sizes ($\Delta \in \{0.05, 0.1, 0.2, 0.3\}$). In both scenarios:

\begin{itemize}
    \item WAPTS shows faster convergence and lower variance.
    \item TS remains unstable or conservative, especially at lower effect sizes.
    \item WAPTS slightly underestimates the true posterior mean in early stages, but maintains a tighter confidence band.
\end{itemize}

Additional results for $N=239$ in the fixed gap setting, which follow the same pattern as Figure~\ref{fig:posterior_change_main}, are provided in Appendix~\ref{appendix:posterior}. To further assess WAPTS in real-world–inspired conditions, we run targeted simulations in three scenarios: (1) a motivational-messaging platform with $N=50$ and $K=2$ treatments; (2) a learnersourcing task with $N=300$ and $K=50$; and (3) an e-mental-health study with $N=1000$ and $K=24$. Full tables and correct-assignment rates are in Appendix~\ref{appendix:simulations}.In more data-sparse settings (larger $K/N$), WAPTS achieves higher average reward and lower posterior estimation error than TS, with improvements that are consistent across replications (see Appendix~\ref{appendix:simulations}).

%% file: sections/8_lenient_regret.tex
\section{Selecting Multiple Good treatments: Educational Value of Lenient Regret}
\label{sec:lenient-educational}

In educational settings—especially with limited student populations and high variability—it is often more practical to identify several \emph{good-enough} treatments rather than commit to a single best one. To support this flexibility, we analyze our policys under a \emph{lenient regret} framework. This perspective is particularly valuable in data-sparse environments: insisting on a unique best can lead to excessive exploration or unstable conclusions. By contrast, selecting a small set of well-performing treatments offers a more robust and scalable policy for adaptive learning.

In our simulations, we set the tolerance to $\epsilon=\eta\cdot\Delta$, where $\Delta$ is the effect size and $\eta\in[0,1]$ controls how forgiving the criterion is (larger $\eta$ widens the “practically equivalent” band). Any treatment whose success probability satisfies $\theta_i \ge \theta^*-\epsilon$ is treated as practically equivalent to the top performer. We mirror the earlier example with $K=10$ treatments and $N=239$ students, and evaluate $\Delta\in\{0.05,\,0.10,\,0.20,\,0.30\}$.

\subsection{Lenient Regret Trends Across Policies}

\begin{figure}[h]
    \centering
    \includegraphics[width=\columnwidth]{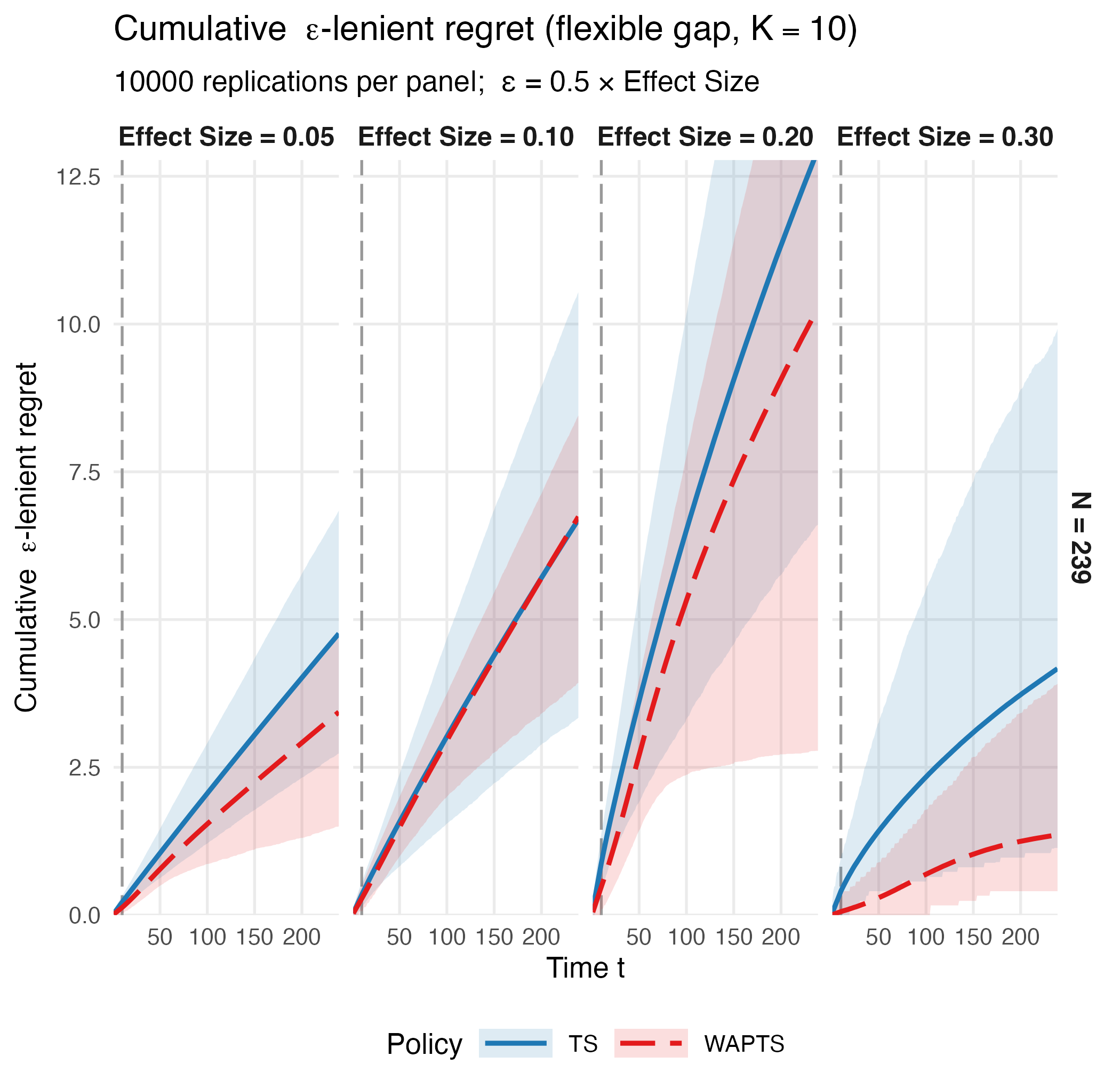}
    \caption{Cumulative $\epsilon$-lenient regret over time under the flexible-gap setting ($K=10$, $N=239$, $\epsilon=0.5\times\Delta$). Lines show Monte Carlo means for TS (blue) and WAPTS (red); ribbons are 95\% bands; the vertical dashed line indicates burn-in.}
    \label{fig:lenient_regret_arms}
\end{figure}

We start by comparing the average lenient regret across various experimental conditions. Note that the lenient regret is generally analyzable in the flexible gap setting, because, if in the fixed gap setting, the single optimal treatment will already surpass the set-up $\epsilon$ that we use and hence only the optimal treatment will be considered 'optimal' even in the lenient regret setting. Below, we analyze how lenient regret evolves with changes in number of students enrolled in and effect sizes. This analysis helps us understand the trade-offs between WAPTS and TS:

As shown in Figure~\ref{fig:lenient_regret_arms}, WAPTS generally accrues less $\epsilon$-lenient regret than TS at small effects ($\Delta\le 0.10$). At $\Delta=0.10$ the two curves nearly coincide: with $\epsilon=0.5\Delta=0.05$, several treatments lie within the lenient band, so both policies spend most pulls on “good-enough” treatments and end up with similar cumulative regret. For larger effects ($\Delta\ge 0.20$), TS’s stronger exploitation reduces regret later in the run and can slightly outperform WAPTS. Shaded ribbons show 95\% simulation bands; the vertical dashed line marks the burn-in period.

We also look at the convergence speed of both policy given the setting that we have between WAPTS and TS, as shown in Appendix section~\ref{sec:lenient-speed}, WAPTS reaches a majority \emph{faster} at small–moderate effects ($0.05$–$0.20$), is roughly tied/slightly slower at $0.30$, and—although both methods always “commit” under a convergence threshold $\tau{=}0.5$—WAPTS yields higher rates of $\epsilon$-lenient commitments at the smaller effects, while TS has the edge at the largest effect in this configuration.

\subsection{Power Trends for Decision Making}

Beyond average outcomes, we also examine how these policies impact statistical inference with the pair wise hypothesis testing framework — i.e., how confidently we can conclude whether a treatment is meaningfully better than others.

% Inference under ε-lenient criterion (tests vs TRUE θ* − ε)
\begin{table}[t]
\centering
\scriptsize
\setlength{\tabcolsep}{6pt}
\renewcommand{\arraystretch}{0.95}
\caption{Power (\%) under $\epsilon$-lenient tests vs the true threshold ($\theta^*-\epsilon$), with BH at $q=0.05$. Settings: $K{=}10$, $N{=}239$, flexible gap, seeds $=1$–$2000$, $\epsilon=0.5\times$Effect Size.}
\label{tab:power_lenient_true}
\begin{tabular}{c cc cc}
\toprule
\multirow{2}{*}{Effect Size} & \multicolumn{2}{c}{Power: $\ge$1 $\epsilon$-lenient treatment} & \multicolumn{2}{c}{Power: true best treatment} \\
 & TS & WAPTS & TS & WAPTS \\
\midrule
0.05 & 3.4 & 6.3 & 1.0 & 1.7 \\
0.10 & 11.1 & 12.6 & 3.7 & 5.2 \\
0.20 & 39.6 & 38.1 & 16.5 & 24.8 \\
0.30 & 78.3 & 74.1 & 43.6 & 57.5 \\
\bottomrule
\end{tabular}
\end{table}

As shown in the Table~\ref{tab:power_lenient_true} , Power is low for small effects ($\Delta\le 0.10$) and rises sharply by $\Delta=0.30$.
At the smallest effect, WAPTS has a slight edge on “detecting at least one” $\epsilon$-lenient treatment (6.3\% vs.\ 3.4\%).
For moderate and large effects, both policies are similar on the “at least one” metric, while WAPTS more often identifies the true best treatment (e.g., 24.8\% vs.\ 16.5\% at $\Delta=0.20$; 57.5\% vs.\ 43.6\% at $\Delta=0.30$).

The results above reflect one setting (\(K{=}10\), \(N{=}239\)) with flexible gaps. They do \emph{not} show WAPTS as uniformly superior: performance varies with the effect size and objective. At small effects, both policies have low power; at larger effects, TS more often identifies the true best treatment, whereas WAPTS tends to select at least one \(\epsilon\)-lenient treatment at similar or faster speeds. In contexts where several near-optimal options are acceptable, the lenient-regret view is practically useful: it prioritizes a small set of comparably effective treatments rather than forcing a single winner. Practitioners should choose \(\epsilon\) and sample size to match whether the goal is “identify \emph{the} best” or “find a set of good-enough options.”

%% file: sections/9_discussion.tex
\section{Discussion and Next Steps}

The WAPTS policy offers a novel mechanism for adaptively adjusting exploration speed in online experiments, especially in educational settings where balancing fast discovery and equitable treatment allocation is critical. Unlike traditional controls such as fixed learning rates or batch sizes, WAPTS dynamically reweights allocation probabilities based on early empirical performance—enabling faster convergence to effective treatments when sample sizes are limited.

While our method is benchmarked against standard Thompson Sampling (TS), the core idea of outcome-sensitive weighting could generalize to a wider family of adaptive policies, including contextual bandits and continuous-outcome models. Our focus on TS stems from its widespread use and strong theoretical properties, but WAPTS enhances its practical utility by prioritizing learning efficiency over asymptotic optimality.

A next step is to explore extensions to continuous outcomes, where the weighting mechanism may need to account for effect size uncertainty rather than binary success/failure counts. Moreover, we recognize that early performance can be noisy, and WAPTS currently updates in a direction that is difficult to reverse. In real-world educational deployments, this could risk prematurely discarding promising treatments due to statistical "bad luck" early on. To mitigate this, we propose a \textbf{reset mechanism}—a periodic reassessment after a minimum sample threshold is reached—to reintroduce arms that may have been unjustly downweighted.

Finally, our framework demonstrates how a Bayesian model can be used to simulate success probabilities, offering an alternative path to evaluate allocation policies used for adaptive experimentation. At the same time, future empirical work should explore integration with real-time user feedback and external domain knowledge~\cite{zhao_cross-domain_2023}.

%% file: sections/appendix.tex
\appendix
\setcounter{secnumdepth}{2} 

\section{Appendix: Regret Analysis of WAPTS Policy}
\label{appendix:regret-wapts}

In this section, we provide a sketch of the regret bound for the WAPTS policy, adapting the classical regret analysis of Thompson Sampling (TS)~\cite{agrawal_analysis_2012}. The analysis illustrates that WAPTS retains similar regret guarantees, up to constant factors, while favoring faster convergence.

\subsection{Setup and Notation}

Consider a $K$-treatmentes bandit problem with Bernoulli rewards. Let $p_k$ denote the true mean of treatment $k$, with $p^* = \max_k p_k$, and define $\Delta_k = p^* - p_k$ as the gap for treatment $k$. At each round $t$, the policy selects treatment $A_t$ and receives reward $Y_{A_t, t} \sim \mathrm{Bern}(p_{A_t})$. The expected cumulative regret after $n$ rounds is:
\[
\mathbb{E}[R(n)] = \sum_{k=1}^K \mathbb{E}[N_k(n)] \Delta_k,
\]
where $N_k(n)$ is the expected number of times treatment $k$ is selected.

\subsection{WAPTS Selection Rule}

WAPTS maintains Beta posteriors for each treatment:
\[
\theta_k \sim \mathrm{Beta}(\alpha_k, \beta_k), \quad \text{with} \quad \alpha_k = S_k + 1,\; \beta_k = F_k + 1,
\]
where $S_k$ and $F_k$ are the cumulative successes and failures. The empirical success rate is:
\[
\hat{p}_k = 
\begin{cases}
    S_k / (S_k + F_k) & \text{if } S_k + F_k > 0, \\
    1 & \text{otherwise.}
\end{cases}
\]
Let the reweighting function be \( f(p) = (1 + p)p \). WAPTS uses:
\[
w_k = f(\hat{p}_k) \cdot \theta_k,
\]
and selects treatment $A_t = \arg\max_k w_k$.

\subsection{Regret Proof Sketch}

We outline the main proof steps:

\begin{enumerate}
    \item \textbf{Burn-in Phase:}  
    For the first $b$ rounds, WAPTS selects treatments uniformly. The regret here is at most $b \cdot \max_k \Delta_k$.

    \item \textbf{Bounding Suboptimal Treatment Pulls:}  
    For each $k \neq k^*$, we bound:
    \[
    \mathbb{E}[N_k(n)] \leq b + \sum_{t=b+1}^n \mathbb{P}(A_t = k).
    \]
    The event $A_t = k$ occurs when $w_k > w_{k^*}$.

    \item \textbf{Posterior and Empirical Concentration:}  
    Once treatment $k$ has been sampled $T_k \gtrsim \frac{C \log n}{\Delta_k^2}$ times, both $\hat{p}_k$ and $\theta_k$ concentrate near $p_k$. Since \( f(p) = (1 + p)p \) is increasing in $p$, the reweighted scores further widen the gap between suboptimal and optimal treatments.

    \item \textbf{Probability of Mistaken Selection:}  
    Using standard concentration inequalities, for large enough $T_k$:
    \[
    \mathbb{P}(w_k > w_{k^*}) \leq \exp(-c T_k \Delta_k^2).
    \]

    \item \textbf{Total Regret Bound:}  
    Summing over $n$ rounds and all suboptimal treatments:
    \[
    \mathbb{E}[R(n)] \leq bK \cdot \max_k \Delta_k + \sum_{k \neq k^*} O\left( \frac{\log n}{\Delta_k} \right).
    \]
    This matches the standard regret order of Thompson Sampling up to constant factors.
\end{enumerate}

\subsection{Remarks}

\begin{itemize}
    \item The reweighting function \( f(p) = (1 + p)p \) accelerates convergence by increasing discrimination among treatments, especially early on.
    \item However, if $f(p)$ is too aggressive, early random fluctuations may overly favor a suboptimal treatment—suggesting the potential value of future mechanisms such as reset or discounting policies.
    \item WAPTS encourages faster exploitation, which is desirable when strong treatments exist, but could be risky in noisy early-stage environments.
\end{itemize}

% \section{Appendix: Lenient Regret Analysis for WAPTS}
% \label{appendix:lenient-regret}

% In many educational or applied settings, allocating to an treatment that is nearly optimal is sufficient. The lenient regret framework quantifies this idea by ignoring regret when the selected treatment's reward is within $\epsilon$ of the optimal.

% \subsection{Definition}

% Let the set of $\epsilon$-optimal treatments be:
% \[
% \mathcal{A}_{\epsilon} = \{k : p^* - p_k \leq \epsilon \}.
% \]
% Then the $\epsilon$-regret (or lenient regret) is:
% \[
% R^{(\epsilon)}(n) = \sum_{t=1}^n \mathbb{E}[(p^* - p_{A_t}) \cdot \mathbf{1}\{A_t \notin \mathcal{A}_\epsilon\}].
% \]

% \subsection{Sketch of Regret Bound}

% As in the classic case, we bound the expected number of times that clearly suboptimal treatments (i.e., $p_k < p^* - \epsilon$) are selected:

% \[
% \mathbb{E}[N_k(n)] \leq b + O\left( \frac{\log n}{(p^* - p_k)^2} \right).
% \]

% So the lenient regret becomes:
% \[
% \mathbb{E}[R^{(\epsilon)}(n)] = \sum_{k: p^* - p_k > \epsilon} \mathbb{E}[N_k(n)] \cdot (p^* - p_k)
% \]
% \[
% \leq bK \cdot \max_{k: p^* - p_k > \epsilon}(p^* - p_k) + \sum_{k: p^* - p_k > \epsilon} O\left( \frac{\log n}{p^* - p_k} \right).
% \]

% \subsection{Implications}

% \begin{itemize}
%     \item The number of treatments incurring regret is reduced to those with gap larger than $\epsilon$, meaning regret can shrink significantly in near-tie settings.
%     \item This is especially helpful for educational treatments or user-facing experiments, where several near-optimal options may offer similar value in practice.
% \end{itemize}

\section{Representative Student Submissions}
\label{appendix:examples}

The following are representative examples of student-generated adaptive experiment designs:

\begin{tcolorbox}[
  colback=gray!10,
  colframe=gray!50,
  title={Student A: Poor Response},
  sharp corners,
  boxrule=1pt
]
\textbf{Student:} \\
"I don't really get why we need to design an adaptive experiment. I'll just use the same old setup. Honestly, this whole idea seems confusing."
\end{tcolorbox}

The filtered set contained 50 examples of varying quality, including both less complex and more sophisticated submissions, such as:

\begin{tcolorbox}[
  colback=blue!5,
  colframe=blue!50,
  title={Student B: Simple Study Strategies Experiment},
  sharp corners,
  boxrule=1pt
]
\textbf{Experiment Design:}\\[1ex]
"Maybe a simple experiment to evaluate three study strategies: reading, summarizing, and practicing. Each student is randomly assigned one of these strategies, and after the exam, they provide binary feedback: 1 if they found the strategy helpful and 0 if not. This discrete outcome allows us to quickly determine which strategy is most effective."
\end{tcolorbox}

\bigskip

\begin{tcolorbox}[
  colback=darkgreen!5,
  colframe=darkgreen!50,
  title={Student C: Complex Learning Materials Experiment},
  sharp corners,
  boxrule=1pt
]
\textbf{Experiment Design:}\\[1ex]
"I design an adaptive experiment to assess restaurant satisfaction on a continuous scale. The treatments remain the same as from the example (three restaurants: A, B, and C), but the outcome is now measured as a continuous satisfaction score ranging from 0 to 100. Additionally, this design incorporates covariates such as prior dining experiences and individual food preferences, and employs a hierarchical Bayesian model for dynamic allocation of participants. This approach allows us not only to determine the best restaurant but to gain a detailed understanding of the factors driving overall customer satisfaction."
\end{tcolorbox}

\section{WAPTS Policy Details}
\label{appendix:wapts}

 Pseudocode for the WAPTS policy used in our experiments—inputs (per-arm success/failure counts, horizon) and the per-round loop that samples per-arm rates, scores arms, plays the top arm, and updates counts.

\begin{algorithm}[h]
\caption{Weighted Allocation Probability Adjusted Thompson Sampling (WAPTS)}
\label{alg:wapts_full}
\begin{algorithmic}[1]
\Require $k_s[k]$: number of successes for treatment $k$
\Require $k_f[k]$: number of failures for treatment $k$
\Require $n$: number of total trials
\Procedure{WAPTS}{$k_s, k_f, n$}
    \For{$t = 1, 2, \dots, n$}
        \For{$k = 1, \dots, K$}
            \State Sample $\tilde{p}_k \sim \mathrm{Beta}(k_s[k]+1, k_f[k]+1)$
            \State $r_k \gets \frac{k_s[k]}{k_s[k] + k_f[k]}$
            \State $\omega_k \gets (1 + r_k)\,r_k\,\tilde{p}_k$
        \EndFor
        \State $k^* \gets \arg\max_k \omega_k$
        \State Observe $y_t \in \{0,1\}$ from treatment $k^*$
        \If{$y_t = 1$}
            \State $k_s[k^*] \gets k_s[k^*] + 1$
        \Else
            \State $k_f[k^*] \gets k_f[k^*] + 1$
        \EndIf
    \EndFor
\EndProcedure
\end{algorithmic}
\end{algorithm}

\section{Additional Posterior Visualizations}
\label{appendix:posterior}

To complement Section~6, we include additional results comparing WAPTS and TS across a sample size of $N=239$ and both in the fixed-gap setting.

\subsection*{Fixed-Gap Posterior Evolution ($N=239$)}

\begin{figure}[H]
    \centering
    \includegraphics[width=0.45\textwidth]{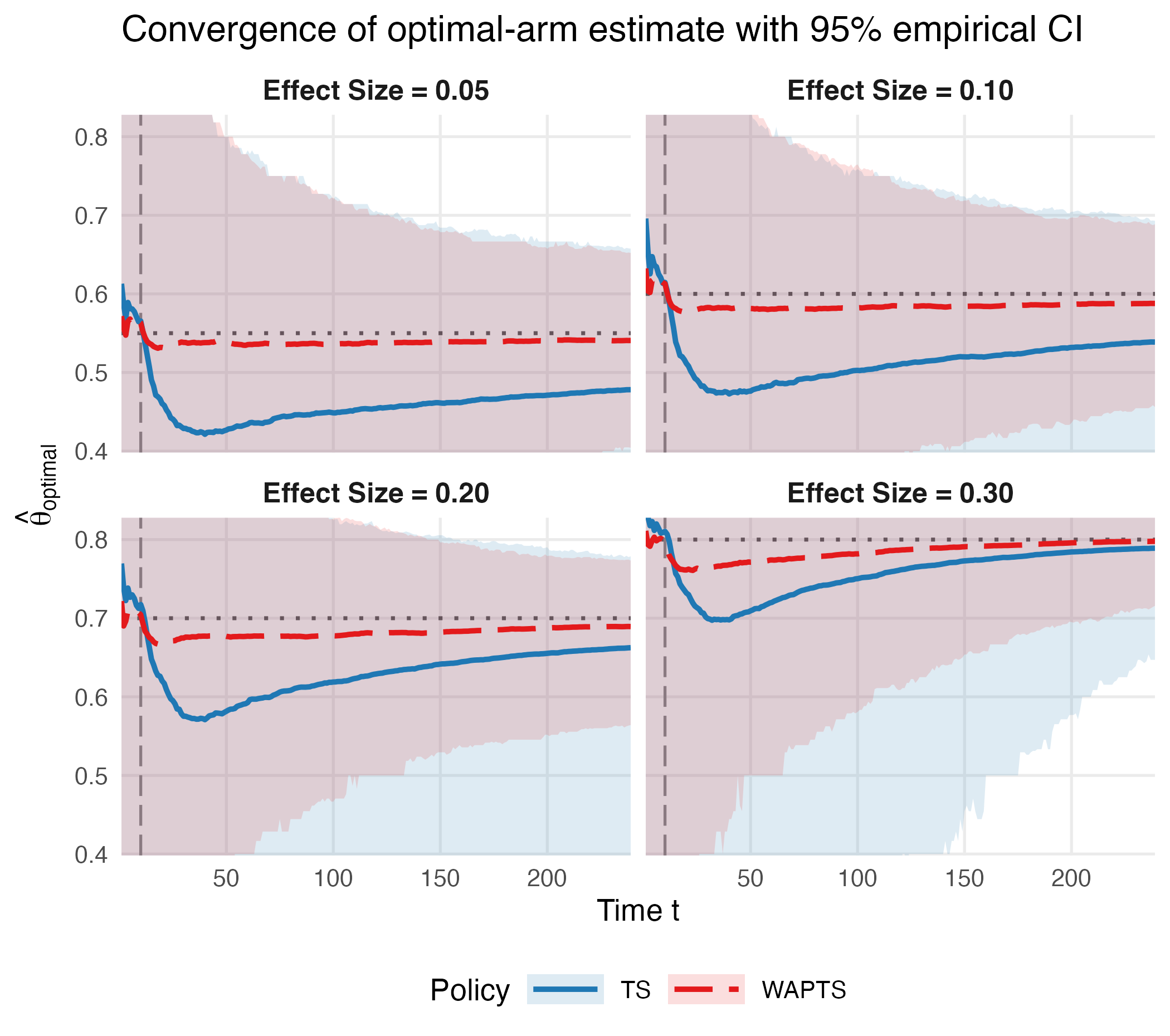}
    \caption{Posterior evolution (Fixed-Gap) with $N = 239$. WAPTS shows tighter estimates and earlier convergence.}
    \label{fig:posterior_change_500_fixed}
\end{figure}

% Optional: Include the full progression plot across all effect sizes
% \subsection*{Posterior Trace Across Effect Sizes}
% \begin{figure}[h]
%     \centering
%     \includegraphics[width=0.45\textwidth]{figures/full_posterior_trace.png}
%     \caption{Posterior mean traces for all treatments across effect sizes (Fixed-Gap).}
%     \label{fig:posterior_trace_all}
% \end{figure}

\section{Appendix: Majority-Commit Metric and Lenient Correctness}
\label{sec:lenient-speed}

\paragraph{Majority-commit time.}

To evaluate the convergence speed and compare WAPTS with TS, we define the following metric to capture and quantify how quickly different allocation policys converge.
Let the assigned treatment be $A_t\in\{1,\dots,K\}$ for rounds $t=1,\dots,N$.
The \emph{majority-commit time} at threshold $\tau\in(0,1]$ is
{\small
\[
T_{\mathrm{maj}}(\tau)=\min\Bigl\{t\ge \text{burn-in}:\ 
\max_{a} \frac{1}{N-t+1}\sum_{s=t}^{N}\mathbf{1}\{A_s=a\}\ge\tau\Bigr\}.
\]
}
We set $\tau=0.5$ so that, in our $K=10$, $N=239$ setting, both TS and WAPTS reach a majority before the horizon; the metric then focuses on \emph{how fast} each policy settles rather than on \emph{whether} it settles.

\paragraph{Lenient correctness.}
Let $\theta_a$ be treatment $a$'s success probability and $\theta^*=\max_j \theta_j$. An treatment $a$ is \emph{$\epsilon$-leniently correct} if $\theta^*-\theta_a \le \epsilon$, with
\[
\epsilon \;=\; \eta \times \text{Effect Size}, \qquad \eta=0.5.
\]

WAPTS reaches a majority \emph{faster} at small–moderate effects ($0.05$–$0.20$), is roughly tied/slightly slower at $0.30$, and—although both methods always “commit” under $\tau{=}0.5$—WAPTS yields higher rates of $\epsilon$-lenient commitments at the smaller effects, while TS has the edge at the largest effect in this configuration.

\begin{table}[H]
\centering
\setlength{\tabcolsep}{3.5pt}      % tighter columns
\renewcommand{\arraystretch}{0.95}  % tighter rows
\scriptsize

% --------- Top: Convergence speed ---------
\begin{minipage}{0.98\columnwidth}
\centering
\caption{Convergence speed (mean$\pm$SD after burn-in, conditional on commit). Gap = flexible; $K{=}10$; $N{=}239$; seeds = 1--2000; $\tau{=}0.5$; $\epsilon=\eta\times$Effect Size, $\eta{=}0.5$.}
\label{tab:speed_maj}
\begin{tabular}{@{}c c c c@{}}
\toprule
ES & TS $\mu\pm\sigma$ & WAPTS $\mu\pm\sigma$ & $\Delta\mu$ (WAPTS$-$TS) \\
\midrule
0.05 & 209.4$\pm$36.0 & 200.0$\pm$59.9 & \textbf{$-9.4$} \\
0.10 & 204.1$\pm$42.6 & 192.9$\pm$67.8 & \textbf{$-11.3$} \\
0.20 & 153.7$\pm$80.8 & 133.8$\pm$97.7 & \textbf{$-20.0$} \\
0.30 & 142.1$\pm$84.2 & 143.4$\pm$96.2 & $+1.3$ \\
\bottomrule
\end{tabular}
\end{minipage}

\vspace{5pt}

% --------- Bottom: Commitment & lenient correctness ---------
\begin{minipage}{0.98\columnwidth}
\centering
\caption{Commitment and $\epsilon$-lenient correctness (same settings as Table~\ref{tab:speed_maj}).}
\label{tab:commit_lenient}
\begin{tabular}{@{}c c c c@{}}
\toprule
ES & \% commit TS/WAPTS & \% lenient TS/WAPTS & $\Delta$ len. (pp) \\
\midrule
0.05 & 100.0 / 100.0 & 33.3 / 51.3 & \textbf{+18.0} \\
0.10 & 100.0 / 100.0 & 67.0 / 80.8 & \textbf{+13.8} \\
0.20 & 100.0 / 100.0 & 82.7 / 78.8 & $-3.9$ \\
0.30 & 100.0 / 100.0 & 95.8 / 81.1 & $-14.7$ \\
\bottomrule
\end{tabular}
\end{minipage}

\vspace{-2mm}
\end{table}

\section{Simulation-Based Analysis of Applications}
\label{appendix:simulations}

In this appendix, we provide simulation results that illustrate how different adaptive policies perform across real-world-inspired settings, with a focus on three application domains: a web-based educational platform, a learnersourcing evaluation, and an e-mental health treatment.

\subsection{Application on a WEB-based Educational Platform ($N =50$)}

In our recent deployment on an online educational platform for university students, we evaluated the effectiveness of two motivational messages aimed at promoting participation in an optional course activity. To better understand the impact of different policy designs, we performed simulations with a sample size of 50 participants and two treatments. Note that the ratio between the sample size and the number of treatments is an important factor in adaptive experiments, as a high ratio facilitates sufficient exploration of each treatment. Under a small effect size setting ($\Delta = 0.1$), our simulation results indicate that WAPTS performs significantly better in terms of average outcomes over 200 simulated trials for both treatment scenarios, with nearly equal outcome deviations. Furthermore, in selecting the correct treatment to exploit, WAPTS outperforms TS and UR by achieving a higher probability of correct selection and a lower deviation.

\begin{table}[H]
\centering
\resizebox{\columnwidth}{!}{%
\begin{tabular}{|l|c|c|c|c|c|}
\hline
\textbf{Policy}   & \textbf{Mean Correct Rate} & \textbf{SD Correct Rate} & \textbf{Mean outcome} & \textbf{SD outcome} & \textbf{WAPTS Superior Count} \\
\hline
TS\_multi         & 0.654                      & 0.241                    & 0.564                & 0.075            & 131                           \\
\hline
UR\_multi         & 0.496                      & 0.068                    & 0.546                & 0.075            & 133                           \\
\hline
WAPTS\_multi      & 0.879                      & 0.149                    & 0.590                & 0.072            & --                           \\
\hline
\end{tabular}%
}
\caption{Comparison of policies: Mean and standard deviation (SD) of correct rates and outcomes for experiments with \(N = 50\) participants and \(K = 2\) treatments under \textbf{Flexible Gap Scenario}. The final column indicates the number of simulations (out of 200) in which WAPTS outperformed the respective policy.}
\label{tab:n_50_k_2_sce1}
\end{table}

\begin{table}[H]
\centering
\resizebox{\columnwidth}{!}{%
\begin{tabular}{|l|c|c|c|c|c|}
\hline
\textbf{Policy} & \textbf{Mean Correct Rate} & \textbf{SD Correct Rate} & \textbf{Mean outcome} & \textbf{SD outcome} & \textbf{WAPTS Superior Count} \\
\hline
TS    & 0.600 & 0.249 & 0.569 & 0.075 & 119 \\
\hline
UR    & 0.496 & 0.068 & 0.551 & 0.074 & 120 \\
\hline
WAPTS & 0.869 & 0.144 & 0.588 & 0.076 & -- \\
\hline
\end{tabular}%
}
\caption{Comparison of policies: Mean and standard deviation (SD) of correct rates and outcomes for experiments with \(N = 50\) participants and \(K = 2\) treatments under \textbf{Fixed Gap Scenario}. The final column indicates the number of simulations (out of 200) in which WAPTS outperformed the respective policy.}
\label{tab:n_50_k_2_sce2}
\end{table}

\subsection{Application of the  Learnersourcing Example ($N = 300$)}

In this part of the study, we examine the learnersourcing scenario within an active recall activity using all three policies. Initially, around 250 examples of concept explanations were collected from students. To ensure quality, a first-round screening was performed using large language models (LLMs) to filter out non-helpful or roughly drafted examples. This screening reduced the pool to 50 better-quality examples, which were then presented to a group of students for evaluation. In the evaluation task, students rated the helpfulness of each example, and the simulation was configured with an effect size of $\Delta = 0.2$ (representing a medium effect).

Tables~\ref{n_300_k_50_sce1} and \ref{n_300_k_50_sce2} summarize the simulation results for two different scenarios. From an average outcome perspective, TS performs slightly better than WAPTS and UR in Flexible Gap Scenario, but worse in Fixed Gap Scenario. However, the differences are minimal considering the number of treatments in the experiment. In contrast, all policies suffer from data sparsity, and as a result, the correct assignment rate is considerably low (around 2\% for TS and UR, and 5\% for WAPTS).

\begin{table}[H]
\centering
\resizebox{\columnwidth}{!}{%
\begin{tabular}{|l|c|c|c|c|c|}
\hline
\textbf{Policy}   & \textbf{Mean Correct Rate} & \textbf{SD Correct Rate} & \textbf{Mean outcome} & \textbf{SD outcome} & \textbf{WAPTS Superior Count} \\
\hline
TS\_multi         & 0.028                      & 0.025                    & 0.617                & 0.029            & 90                           \\
\hline
UR\_multi         & 0.021                      & 0.008                    & 0.604                & 0.027            & 111                          \\
\hline
WAPTS\_multi      & 0.051                      & 0.013                    & 0.611                & 0.030            & --                           \\
\hline
\end{tabular}%
}
\caption{Comparison of policies: Mean and standard deviation (SD) of correct rates and outcomes for experiments with \(N = 300\) participants and \(K = 50\) treatments under \textbf{Flexible Gap Scenario}. The final column indicates the number of simulations (out of 200) in which WAPTS outperformed the respective policy.}
\label{n_300_k_50_sce1}
\end{table}

\begin{table}[H]
\centering
\resizebox{\columnwidth}{!}{%
\begin{tabular}{|l|c|c|c|c|c|}
\hline
\textbf{Policy} & \textbf{Mean Correct Rate} & \textbf{SD Correct Rate} & \textbf{Mean outcome} & \textbf{SD outcome} & \textbf{WAPTS Superior Count} \\
\hline
TS    & 0.048 & 0.036 & 0.506 & 0.029 & 116 \\
\hline
UR    & 0.021 & 0.008 & 0.505 & 0.027 & 113 \\
\hline
WAPTS & 0.048 & 0.014 & 0.515 & 0.031 & -- \\
\hline
\end{tabular}%
}
\caption{Comparison of policies: Mean and standard deviation (SD) of correct rates and outcomes for experiments with \(N = 300\) participants and \(K = 50\) treatments under \textbf{Fixed Gap Scenario}. The final column indicates the number of simulations (out of 200) in which WAPTS outperformed the respective policy.}
\label{n_300_k_50_sce2}
\end{table}

\subsection{Application of an E-Mental Health Project ($N =1000$)}

Another example involves improving students' mental well-being through a multi-week interactive text messaging program. One of the primary objectives was to identify the optimal times for sending messages to maximize timely user responses. The experiment employed 24 treatment treatments—each corresponding to a different offset hour from the user's permissible contact time—with a binary outcome variable for user response.

After the deployment phase, we replicated the experimental conditions in simulations involving 1,000 participants and 24 treatments, with a small effect size ($\Delta = 0.1$). Our simulation results (see Tables~\ref{n_1000_k_24_sce1} and \ref{n_1000_k_24_sce2}) indicate that WAPTS and TS achieved nearly equal average outcomes in both scenarios, outperforming UR. Furthermore, WAPTS showed about 15\% better correct selection of the optimal treatment in both scenarios compared to TS.

\begin{table}[H]
\centering
\resizebox{\columnwidth}{!}{%
\begin{tabular}{|l|c|c|c|c|c|}
\hline
\textbf{Policy}   & \textbf{Mean Correct Rate} & \textbf{SD Correct Rate} & \textbf{Mean outcome} & \textbf{SD outcome} & \textbf{WAPTS Superior Count} \\
\hline
TS\_multi         & 0.063                      & 0.048                    & 0.565                & 0.016            & 104                           \\
\hline
UR\_multi         & 0.042                      & 0.006                    & 0.560                & 0.015            & 124                           \\
\hline
WAPTS\_multi      & 0.097                      & 0.038                    & 0.565                & 0.015            & --                           \\
\hline
\end{tabular}%
}
\caption{Comparison of policies: Mean and standard deviation (SD) of correct rates and outcomes for experiments with \(N = 1000\) participants and \(K = 24\) treatments under \textbf{Flexible Gap Scenario}. The final column indicates the number of simulations (out of 200) in which WAPTS outperformed the respective policy.}
\label{n_1000_k_24_sce1}
\end{table}

\begin{table}[H]
\centering
\resizebox{\columnwidth}{!}{%
\begin{tabular}{|l|c|c|c|c|c|}
\hline
\textbf{Policy} & \textbf{Mean Correct Rate} & \textbf{SD Correct Rate} & \textbf{Mean outcome} & \textbf{SD outcome} & \textbf{WAPTS Superior Count} \\
\hline
TS    & 0.114 & 0.076 & 0.512 & 0.018 & 102 \\
\hline
UR    & 0.042 & 0.006 & 0.503 & 0.015 & 131 \\
\hline
WAPTS & 0.129 & 0.082 & 0.512 & 0.017 & -- \\
\hline
\end{tabular}%
}
\caption{Comparison of policies: Mean and standard deviation (SD) of correct rates and outcomes for experiments with \(N = 1000\) participants and \(K = 24\) treatments under \textbf{Fixed Gap Scenario}. The final column indicates the number of simulations (out of 200) in which WAPTS outperformed the respective policy.}
\label{n_1000_k_24_sce2}
\end{table}